\documentclass[11pt,a4]{article}
\usepackage{times,fancyhdr}
\usepackage[dvips]{graphicx}
\usepackage{amsmath}
\usepackage{amssymb}
\usepackage{array}
\usepackage[small,bf]{caption}
\usepackage{color}
\usepackage[yyyymmdd,hhmmss]{datetime}
\usepackage{fancyhdr}
\usepackage{fancyvrb}
\usepackage{hyperref}
\usepackage{longtable}
\usepackage{natbib}
\usepackage{setspace}
\usepackage{framed}
\usepackage{titlesec}
\usepackage{titling}
\usepackage{xcolor}
\usepackage{Verbatim,framed}

\fancyheadoffset[R]{0pt}
\fancyfootoffset[R]{0pt}
\definecolor{orange}{rgb}{1.0,0.5,0.0}
\definecolor{aqgr}  {rgb}{0.0,1.0,0.6} 
\definecolor{viol}  {rgb}{0.8,0.6,0.8}
\definecolor{figdr} {rgb}{1.0,1.0,1.0} 
\definecolor{colne} {rgb}{1.0,0.0,1.0} 
\definecolor{coldr} {rgb}{1.0,0.8,0.0} 
\definecolor{colop} {rgb}{0.0,1.0,1.0} 
\definecolor{colok} {rgb}{0.7,1.0,0.7} 

\newcommand{\sbo}{\scalebox{1.05}}

\newcommand\pardr[1]{\colorbox{coldr}{\textit{#1}}}

\newcommand\parok[1]{\colorbox{colok}{\textbf{#1}}}
\newcolumntype{P}[1]{>{\centering\arraybackslash}p{#1}}
\def\afhead{0}

\title{\vspace{0.0cm}\bfseries{\textsc{
   And/or trade-off in artificial neurons: \\ 
   impact on adversarial robustness}}}

\author{\normalsize \textbf{Alessandro Fontana} \\ 
\texttt{\normalsize fontalex00@gmail.com} 
\vspace{-1.0cm}} \date{}
\begin{document}

\pretitle{%
\begin{center}\LARGE
\vskip -2.2cm
\rule{\textwidth}{2.0pt}
\par
\vskip 0.5cm
}
\posttitle{
\par
\rule{\textwidth}{2.0pt}
\end{center}
\vskip -0.0cm
}

\maketitle
   
\vspace*{-0.5cm}
\begin{abstract}
\if\afhead1 {\parok{[xxxx]}} \fi
Despite the success of neural networks, the issue of classification robustness remains, particularly highlighted by adversarial examples. In this paper, we address this challenge by focusing on the continuum of functions implemented in artificial neurons, ranging from pure AND gates to pure OR gates. Our hypothesis is that the presence of a sufficient number of OR-like neurons in a network can lead to classification brittleness and increased vulnerability to adversarial attacks. We define AND-like neurons and propose measures to increase their proportion in the network. These measures involve rescaling inputs to the [-1,1] interval and reducing the number of points in the steepest section of the sigmoidal activation function. A crucial component of our method is the comparison between a neuron's output distribution when fed with the actual dataset and a randomized version called the ``scrambled dataset.'' Experimental results on the MNIST dataset suggest that our approach holds promise as a direction for further exploration.
\end{abstract}

\section{The issue of robustness}  

\if\afhead1 {\parok{adversarial examples}} \fi
Around 2013, neural networks achieved human-level performance in image classification, sparking the emergence of deep learning. However, accuracy alone did not satisfy researchers, prompting investigations into the robustness and interpretability of these models. To shed light on these aspects, \citep{Szegedy13} presented two intriguing findings. The first pertained to information storage within networks, while the second instantly captured the scientific community's attention. The second finding revealed that by introducing imperceptible, artificially crafted noise to a correctly classified image, a highly probable misclassification could be induced.

\if\afhead1 {\parok{related problems: interpretability}} \fi
This problem is closely linked to the challenge of understanding how information is encoded in neural networks \citep{Samek17}. In supervised learning using stochastic gradient descent, hidden neurons learn features as a means to minimise classification errors in the output layer. However, the statistical properties of these intermediate neuron features and their impact on the network's classification performance remain poorly understood. The study of adversarial examples has evolved into a distinct field, currently engaged in an ongoing arms race where attackers maintain the upper hand. Every defense proposed is met with new methods and attacks emerging on a weekly basis \citep{Yuan18}.

\if\afhead1 {\parok{some defences}} \fi
In the pursuit of enhancing the robustness of neural networks, researchers have proposed various approaches. One such technique, presented by \citep{Papernot17}, is defensive distillation. This method aims to mitigate the network's tendency to overfit the training data by encouraging it to generalise better. While defensive distillation has demonstrated resistance to certain types of attacks, it may be more vulnerable to others. Another approach, outlined in \citep{Metzen17}, involves training a secondary model on the internal representations of the neural network. This auxiliary model aims to predict whether an input example is adversarial or benign. By leveraging the learned internal representations, this approach seeks to improve the network's ability to detect and defend against adversarial examples.

\if\afhead1 {\parok{adversarial train}} \fi
The current state-of-the-art defence, called adversarial training, consists in adding adversarial examples to the training set. One of the best approaches along this direction, proposed in \citep{Madry17, Ilyas19}, generates adversarial examples within the training cycle. This method produces models more robust to adversarial perturbations, but requires knowledge of all the possible attacks which might occur, and in some cases breaks down completely. Moreover, it can considerably increase training times, although countermeasures to address this issue have already been proposed \citep{Shafahi19}.

\if\afhead1 {\parok{certified robustness}} \fi
One line of research aims at escaping this cat and mouse game between attackers and defenders, by providing evidence of ``certified robustness''. This certification guarantees that, under certain conditions, the network performance is stable. Using this approach, \citep{Carlini22} were able to achieve state-of-the-art certified adversarial robustness (71\% accuracy on ImageNet) under adversarial perturbations constrained to be within an L2-norm of 0.5, by relying exclusively on off-the-shelf pretrained models.  

\if\afhead1 {\parok{saturation defence}} \fi
In the quest for robust defences against adversarial attacks, researchers have explored the realm of biological-inspired solutions. One such approach, presented in \citep{Nayebi17}, proposes pushing network neurons to operate within the saturated region of the activation curve. This entails directing ReLU neurons to the negative input space and leveraging the flat part of the sigmoid curve. Saturating networks exhibit a significantly larger positive excess kurtosis in the weight distribution when compared to conventional networks. This property, inspired by biological systems, held promise as a defence mechanism. However, despite its initial appeal, this defence has also been found vulnerable to attacks, as documented in \citep{Brendel17}. 

\if\afhead1 {\parok{Noise based-defences}} \fi
The defence we are going to propose makes use of random noise to increase the robustness of neural networks to perturbations. Among the works along this line of research, the idea of ``randomized smoothing'' appears particularly promising \citep{Bai19, Cohen19}. Another interesting proposal \citep{Webb19} makes use of a statistical technique to estimate the probability of rare events, such as those associated to the occurrence of adversarial examples. This approach, although lacking theoretical certification, has the advantage of providing an estimate of the probability of occurrence. 

\if\afhead1 {\parok{paper struct}} \fi
This paper is divided into six parts: this first section is the introduction; the second section deals with the AND/OR trade-off in artificial neurons; the third section presents an auxiliary tool called Scrambled Dataset; the fourth section introduces our proposed defence technique; the fifth section presents some experimental results; the last section draws the conclusions and outlines future research directions.

\section{AND/OR trade-off in artificial neurons}  
\label{section:trade-off}

\begin{figure}[t] \begin{center} \hspace*{-0.0cm}
\includegraphics[width=17.00cm]{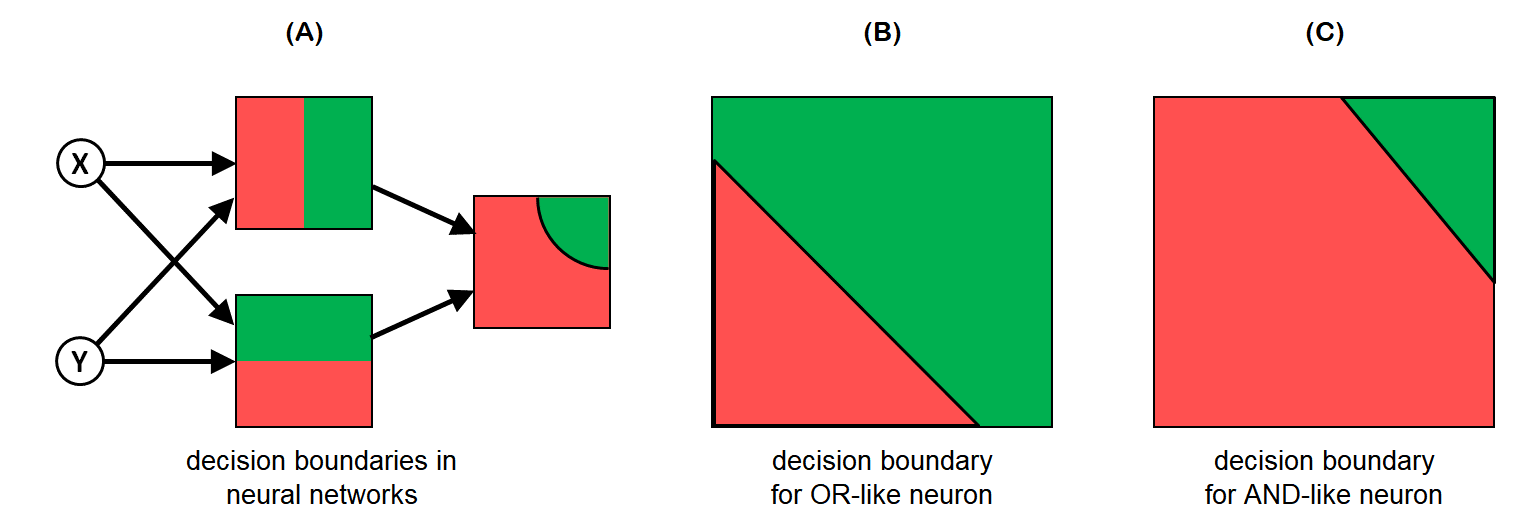}
\caption{Neuron decision boundaries. The decision boundary in panel B corresponds to equation $x+y-0.9=0$, which implements an OR function. The decision boundary in panel C corresponds to equation $x+y-1.8=0$, which implements an AND function. In the case of OR-like neurons, a larger number of input value combinations (proportional to the green area) triggers neuron activation.}
\label{decisionbound}
\end{center} \end{figure}

\if\afhead1 {\parok{neuron function}} \fi
Mathematically, an artificial neuron performs a binary partition of the input space by means of a hyperplane defined by the weight and bias values: the function implemented by the neuron can take different forms depending on such values. To fix the ideas, let us analyse a neural network in which the neurons' inputs and outputs can take only two values (0 and 1), and the neuron implements a function defined by $S(\sum_i (w_i \cdot x_i) + b) = S(\textbf{w} \cdot \textbf{x} + \textbf{b})$ (where S is the sigmoid function). 

\if\afhead1 {\parok{AND/OR trade-off, static}} \fi
Let us consider a first neuron N1 (Fig.~\ref{decisionbound}, panel B) with two inputs with weight = 1 and bias = $-0.9$ (all the other weights are = 0). This neuron fires when one of the inputs or both are = 1: therefore, this neuron implements an OR function. Let us consider a second neuron N2 (Fig.~\ref{decisionbound}, panel C) with two inputs with weight = 1, but with bias = $-1.8$. In order for this neuron to fire, both inputs must be equal to 1: therefore, this neuron implements an AND function. These two cases can be considered as two extremes of a continuum, with many possible intermediate functions in between.

\if\afhead1 {\parok{AND/OR trade-off, dynamic}} \fi
This characterisation of the AND/OR trade-off is based on the neuron's parameters (weights and bias), and can be called \textit{static}. An AND/OR trade-off can be seen also in the statistical behaviour of the neuron input values, a perspective that we might call \textit{dynamic}. In order to get a better intuition of this phenomenon, let us call N the number of positively weighted inputs of a neuron (all with the same weight value for simplicity) and K the number of positively weighted inputs which must be equal to 1 for the neuron to get activated. The binomial coefficient (N,K) is the number of input combinations that lead to neuron activation. For example, if a neuron has 5 positively weighted inputs and 3 are needed for neuron activation, the binomial coefficient gives $(5,3) = 5 \cdot 4 \cdot 3$. This corresponds to the number of combinations of active inputs needed for neuron activation: (1,2,3), (1,2,4), (1,2,5), (2,3,4), etc. From what we have said, it seems that AND-like functions are characterised by a lower number of input combinations needed for activation, which is achieved when N is low and when K is close to N. Intuitively, AND-like neurons capture information in a more precise way, as they require the simultaneous occurrence of values in a well-defined set of inputs. 

\if\afhead1 {\parok{hypothesis on adversarial examples}} \fi
We propose a hypothesis suggesting that the prevalence of adversarial examples in neural networks could be attributed, at least in part, to the presence of a substantial number of OR-like neurons. These neurons possess a multitude of inputs that contribute incrementally to the computational process, with each input making a small individual impact. Consequently, a wide range of input value combinations can activate these neurons. While this characteristic may facilitate network training, it also renders the network susceptible to small perturbations that affect numerous neurons. These perturbations, often imperceptible or irrelevant to human perception, have the potential to alter the decision boundaries of many downstream neurons, ultimately leading to the generation of adversarial examples.

\section{Auxiliary tool: Scrambled Dataset}  

\begin{figure*}[t] \begin{center} \hspace*{-0.25cm}
\includegraphics[width=17.50cm]{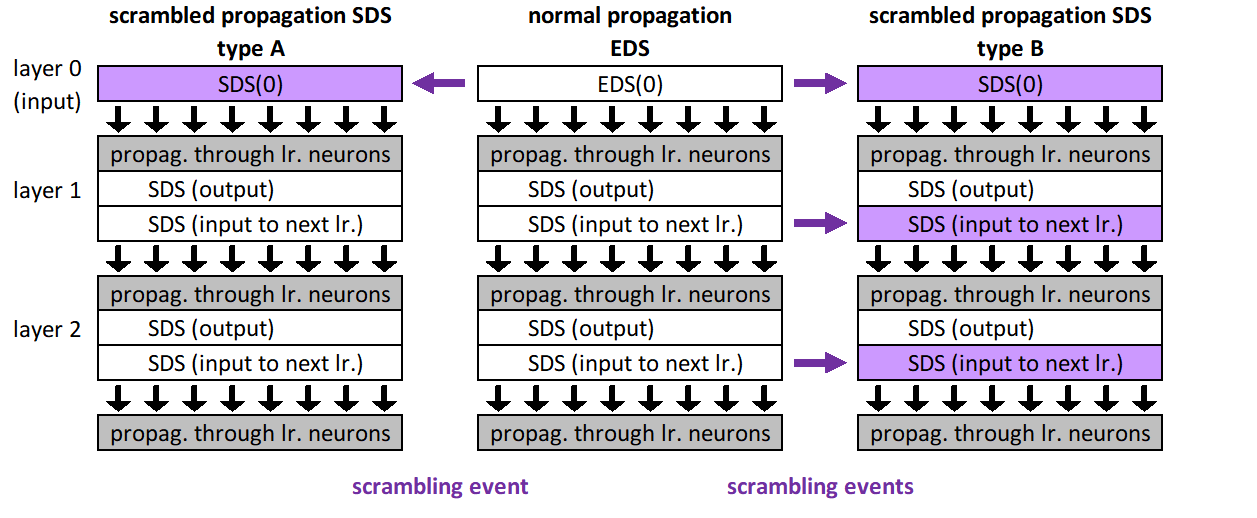}
\caption{
\if\afhead1 {\parok{figurex}} \fi
\if\afhead1 {\parok{kaption}} \fi
Normal and scrambled propagation. The central panel shows the normal propagation of the signal through the layers of a neural network. From the normal propagation, two types of scrambled propagation can be derived. In type A, only layer 0 (EDS(0)) is scrambled. The obtained signal is then propagated through the network as in the normal case. In type B, instead, each layer is scrambled, i.e. it is computed by scrambling the values of the layer obtained from the normal propagation. For example, SDS(1) is computed by scrambling EDS(0) and propagating the scrambled signal through the neurons of the first layer, SDS(2) is computed by scrambling EDS(1) and propagating the scrambled signal through the neurons of the second layer, and so on.}
\label{propagation}
\end{center} \end{figure*}

\begin{figure*}[t] \begin{center} \hspace*{-0.0cm}
\includegraphics[width=17.00cm]{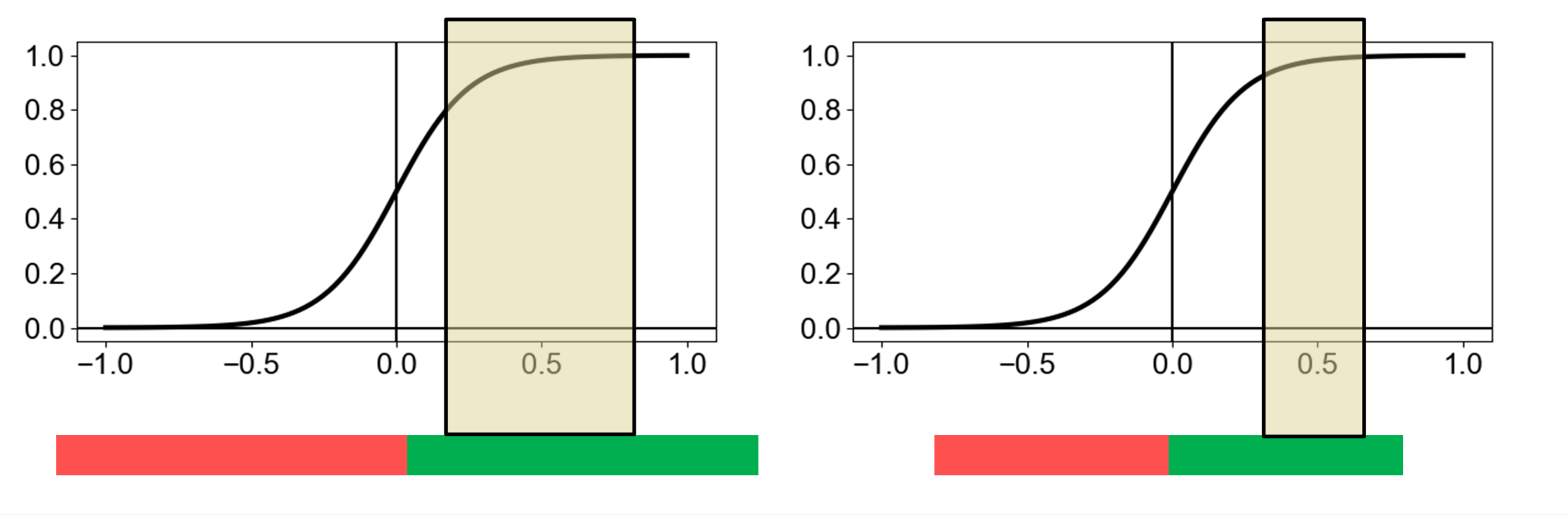}
\caption{
\if\afhead1 {\parok{figurex}} \fi
\if\afhead1 {\parok{kaption}} \fi
Effect of the L1 norm of the vector of a neuron's positive weights (P) on resilience to perturbations. \textbf{Left:} If P (represented by the size of the green bar) is high, even a perturbation corresponding to a low fraction (30\%) of P is sufficient to cause a big displacement of the neuron's operating point. \textbf{Right:} This does not happen if P is lower: in this case, the same \% amount of perturbation produces a more limited displacement of the neuron's operating point. Analogous considerations can be made for the vector of negative weights.}
\label{whtbudget}
\end{center} \end{figure*}

\if\afhead1 {\parok{definition of SDS}} \fi
Given a neural network, let us introduce the notion of \textit{Extended Dataset} (EDS), defined as the union, for all examples, of all values of dataset variables and successive layers' neurons: EDS(lr)(nr)(ex) (lr = layer number, nr = neuron number, ex = example number; the element EDS(0)(nr)(ex) corresponds to the dataset). Secondly, we define another, artificial dataset called \textit{Scrambled Dataset} (SDS), constructed by drawing values randomly from EDS. Here the algorithm's pseudocode for a layer lr (Rnd(N) returns a random number in the [0,N-1] interval). 

\if\afhead1 {\parok{pseudocode}} \fi
\begin{Verbatim}[baselinestretch=1.0]      
   For ex = 0 To Sizeof_SDS-1
      For nr = 0 To NofNeurons(lr)-1
         ee = Rnd(Sizeof_EDS)
         SDS(lr)(nr)(ex) = EDS(lr)(nr)(ee)
      Next nr
   Next ex
\end{Verbatim}

\if\afhead1 {\parok{two different types of SDS}} \fi
From the ``normal'' signal propagation represented by EDS, two types of scrambled propagations can be derived (Fig.~\ref{propagation}). In type A, only layer 0 (EDS(0)) is scrambled. The obtained signal is then propagated through the network as in the normal case. In type B, instead, each layer is scrambled, i.e. it is computed by scrambling the values of the layer obtained from the normal propagation. For example, SDS(1) is computed by scrambling EDS(0) and propagating the scrambled signal through the neurons of the first layer, SDS(2) is computed by scrambling EDS(1) and propagating the scrambled signal through the neurons of the second layer, and so on. 

\if\afhead1 {\parok{correlation measures}} \fi
In practise, SDS is a version of EDS in which the single variables have the same marginal probability distribution as in EDS, but where the correlations between variables are ``broken''. It can be used to obtain a generalisation of standard measures used in statistics to capture correlation among variables. Most of these measures are based on some kind of comparison between the joint probability of the variables and the product of their marginal probabilities. One such measure is represented by Pearson correlation coefficient which, in the case of two variables $x_1$ and $x_2$, can be written as: 

\vskip 0.25cm
\begin{tabular}{p{14.6cm} c}
\newcommand{\aden}{{\sigma_{x_1} \cdot \sigma_{x_2}}}
\newcommand{\bfen}{{E[x_1 \cdot x_2] - E[x_1] \cdot E[x_2]}}
\sbo{$ \rho_{x_1,x_2} = \dfrac {cov(x_1,x_2)} \aden = \dfrac \bfen \aden $} & 
\sbo{(1)}
\end{tabular}
\vskip 0.25cm

\if\afhead1 {\parok{cor coef limitations}} \fi
The correlation coefficient has an intrinsic limit: it is only able to capture correlation between two variables. The concept can be extended to an arbitrary number of variables through the covariance matrix, which provides the covariances between all possible couples of variables (matrix element (i,j) holds the covariance between variables i and j): however, this is not a measure of correlation among the variables \textit{considered together}. Given, e.g., three variables $x_1$, $x_2$ and $x_3$, thanks to the covariance matrix one can measure the correlation between variables $x_1$ and $x_2$, between variables $x_1$ and $x_3$, and between variables $x_2$ and $x_3$; not among all variables $x_1$, $x_2$ and $x_3$.

\if\afhead1 {\parok{reformulation}} \fi
However, thanks to the notion of SDS, the product of expected values $E[x_1] \cdot E[x_2]$ in equation (1) can be rewritten as $E[x_1 \cdot x_2]|SDS$ (if the two variables $x_1$ and $x_2$ are uncorrelated, as it is the case in SDS, the product of the expectations is equal to the expectation of the product). Therefore, equation (1) can be reformulated as:

\vskip 0.25cm
\begin{tabular}{p{14.6cm} c}
\newcommand{\aden}{{\sigma_{x_1} \cdot \sigma_{x_2}}}
\sbo{$ \rho_{x_1,x_2} = \dfrac{ 
E[x_1 \cdot x_2]|\mathrm{EDS} - 
E[x_1 \cdot x_2]|\mathrm{SDS}} \aden $} &
\sbo{(2)}
\end{tabular}
\vskip 0.25cm 

\if\afhead1 {\parok{comparison generalisation}} \fi
The next step towards generalising equation (1) consists in replacing variables $x_1$ and $x_2$ with a vector of variables $\textbf{x} = x_{1}, x_{2}, ... , x_{n}$ and replacing function $x_1 \cdot x_2$ with an arbitrary function $F(\textbf{x})$ of the $x_i$. Finally, we replace the comparison implemented by the minus sign with a generic comparison operator, represented by function G, between the distribution D of $F(\textbf{x})$ computed when F is fed with EDS and the distribution D of $F(\textbf{x})$ computed when F is fed with SDS: 

\vskip 0.25cm
\begin{tabular}{p{14.6cm} c}
\sbo{$ cor(\textbf{x}) = G(
D(F(\textbf{x}))|\mathrm{EDS}, 
D(F(\textbf{x}))|\mathrm{SDS}) $} &
\sbo{(3)}
\end{tabular}
\vskip 0.25cm

\if\afhead1 {\parok{contrastive learn}} \fi
The idea of contrasting signal with noise has already inspired many training algorithms. An example is given by Noise Contrastive Estimation (NCE), a method used for unsupervised learning, or feature extraction \citep{Gutmann12, Ceylan18}, that was tested on different problems with good results \citep{vandenOord18}. The main difference between our method and NCE is that, while in the latter noise is only generated at input layer and then propagated through the network, in our case (scrambled propagation type B) the randomised version of the signal is generated at each layer. This provides a local training signal to the layer's neurons, that is independent of the overall network architecture. Similar approaches have been proposed in \citep{Hinton23} and \citep{Chen20} to train neural networks without the need of backpropagation, or in addition to it.


\section{Defence measures} 

\subsection*{The issue with saturation defence} 

\begin{figure}[t] \begin{center} \hspace*{-0.0cm}
\includegraphics[width=17.00cm]{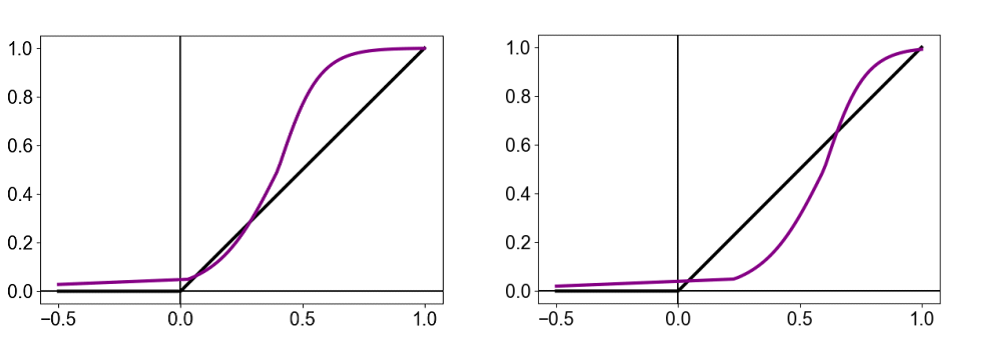}
\caption{
\if\afhead1 {\parok{figurex}} \fi
\if\afhead1 {\parok{kaption}} \fi
AND/OR trade-off in activation function. \textbf{Left:} We say that a neuron activation function is of OR type, if it has high values also for low values of $nics$ (on x axis, the $nics$ value; on y axis, the value of the activation function). \textbf{Right:} We say it is of AND type, if it has high values only for high values (close to 1) of $nics$. The ReLU function is reported for reference.}
\label{andorfunc}
\end{center} \end{figure}

\begin{figure}[t] \begin{center} \hspace*{-0.0cm}
\includegraphics[width=17.00cm]{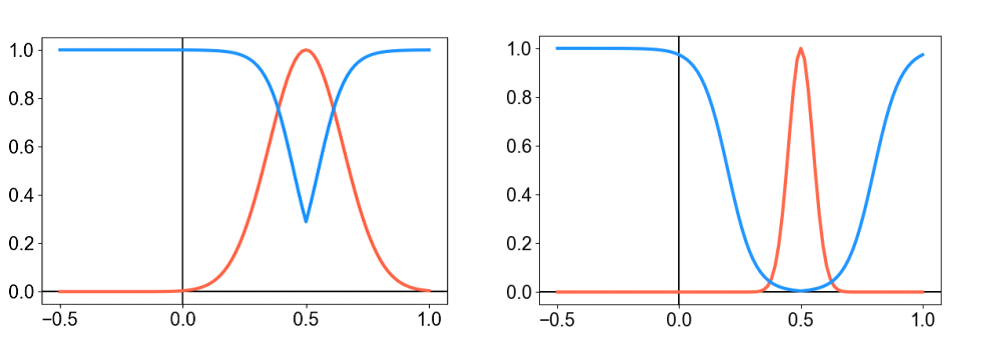}
\caption{
\if\afhead1 {\parok{figurex}} \fi
\if\afhead1 {\parok{kaption}} \fi
AND/OR trade-off in input statistical properties. \textbf{Left:} Expected probability density function (PDF) shapes of a neuron's $nics$ when fed with SDS (red line) and EDS (blue line) for an input behaviour more OR-like. The x-axis represents $nics$ values and the y-axis represents PDF values. The EDS PDF typically exhibits a bimodal distribution, while the SDS PDF follows a Normal distribution. \textbf{Right:} The same PDFs for an input behaviour more AND-like. The SDS PDF appears narrower, and the EDS PDF exhibits a higher probability mass at the extremes. These PDF illustrations are qualitative and do not sum up to 1.}
\label{andorstat}
\end{center} \end{figure}

\if\afhead1 {\parok{build upon saturation defence}} \fi
Our defence proposal builds upon the \emph{saturation defence} proposed in \citep{Nayebi17}. The training scheme developed by the authors, which is inspired by biophysical principles underpinning dendritic computation in neural circuits, generates highly nonlinear, saturated neural networks. The idea is indeed appealing, especially when a sigmoid activation function is used. The central zone of the sigmoid function is characterised by high first derivative values: as a result, input variations are translated to higher output variations and input perturbations are more susceptible to get amplified. If the neuron activation values are close to either 0 or 1, a stronger perturbation is needed to cause a shift in the decision boundary. 

\if\afhead1 {\parok{gradient inactivation and reactivation}} \fi
Unfortunately, the good results reported by the authors seem to have been produced by an unintended side-effect of the procedure. In the saturated regions of the activation function, the gradients of the loss function (used to compute adversarial perturbations) vanish: as a result, the calculation of adversarial examples becomes impossible. However, a method proposed in \citep{Brendel17} was able to resuscitate the gradients and break the defence. Our proposal will try to remedy the aforementioned shortcomings and preserve the good results in terms of adversarial resilience, without the need to have null gradients.

\if\afhead1 {\parok{dependence on L1 norm of weight vector}} \fi
Let us indicate with \textbf{p} and \textbf{n} the vectors of the positive ($p_i$) and negative ($n_i$) weights of a neuron, and $P$ and $N$ the relevant L1 norms: $P = ||\textbf{p}||_1$, $N = ||\textbf{n}||_1$. The problem of the saturation defence is that it critically depends on the value of $P$ and $N$. If $P$ is high, even a perturbation corresponding to a relatively low fraction (30\%) of $P$ is sufficient to cause a large displacement of the neuron's operating point (Fig.~\ref{whtbudget}, left). On the other hand, if $P$ is lower (Fig.~\ref{whtbudget}, right), the same percentage amount of perturbation will produce a more limited displacement of the neuron's operating point, less likely to produce a change in the classification result (analogous considerations can be made for the negative weights).  

\if\afhead1 {\parok{nics, AND/OR tradeoff with actifunc}} \fi
These considerations can be used to obtain a characterisation of the AND/OR nature of a neuron. Let us assume that $P$ (L1 norm of the vector of the neuron positive weights) and $N$ (L1 norm of the vector of the neuron negative weights) are both $\leq 1$. Let us consider the dot product of the weight vector $\textbf{w}$ and the input vector $\textbf{x}$. We call this quantity \textit{neuron input coactivation score}: $nics = \textbf{w} \cdot \textbf{x}$. Thanks to the assumption on the L1 normalisation, $nics$ is always comprised in the [-1,1] interval. With this assumption, we say that a neuron activation function is of OR type, if it has high values also for low values of $nics$ (Fig.~\ref{andorfunc}, left); we say it is of AND type, if it has high values only for high values (close to 1) of $nics$ (Fig.~\ref{andorfunc}, right).  

\if\afhead1 {\parok{AND/OR tradeoff with statistical distribution}} \fi
As already pointed out, the characterisation of the AND/OR trade-off can be observed also in the statistical distribution of the neuron input values. If the inputs of a neuron are correlated, $nics$ will be either very high (when many inputs are simultaneously active) or very low (when many inputs are simultaneously inactive): as a result the PDF will tend to be bimodal (Fig ~\ref{andorstat}, left, blue line). When the neuron is fed with SDS, on the other hand, the PDF will tend to assume a Normal shape (Fig ~\ref{andorstat}, left, red line). These two tendencies will be increased the more the neuron's input displays an AND-like behaviour (Fig ~\ref{andorstat}, right): the peaks of the PDF corresponding to EDS will tend to be farther apart, the variance of the PDF corresponding to EDS will tend to be lower.   

\if\afhead1 {\parok{measures: introduction}} \fi
To promote the development of neurons exhibiting a more AND-like behaviour within the network, we propose the implementation of multiple measures that are specific to each neuron and solely rely on locally available information at the neuron-level. These measures will be integrated into a secondary loss function, which will be combined with the existing loss function employed for the supervised learning task, specifically classification. By incorporating these additional measures into the learning process, we aim to encourage the emergence of neurons that demonstrate enhanced AND-like characteristics within the network architecture.

\subsection*{Weight vector L1 normalisation} 

\if\afhead1 {\parok{weight normalisation}} \fi
The first measure is weight normalisation. This is obtained by imposing that $P$ (L1 norm of vector of positive weights) and $N$ (L1 norm of the vector of negative weights) are both $\leq 1$. As a result, the neuron's input to the activation function is always comprised in the [-1,1] interval.

\vskip 0.25cm
\begin{tabular}{p{7cm} p{7cm} c}
\sbo{$p_i \leftarrow p_i / P$, if $P > 1$} &
\sbo{$n_i \leftarrow n_i / N$, if $N > 1$} &
\sbo{(4)}
\end{tabular}
\vskip 0.25cm

\subsection*{AND-type activation function} 

\if\afhead1 {\parok{sigmoid af}} \fi
The second measure to foster the AND-like behaviour of neurons consists in using an AND-type activation function (Fig.~\ref{andorfunc}, right). This processing step reduces the contribution of lower input values and enhances the contribution of higher input values. This tends to favour the emergence of neurons that, when active, produce high activation values; on the contrary, neurons producing low activation values have little effect on neurons downstream. The application of a sigmoid activation function has the desired effect only in combination with the first measure proposed: in fact, only in this case the $nics$ is equal to 1 if and only if all inputs are simultaneously active. 

\subsection*{Hyper-saturation} 

\if\afhead1 {\parok{saturation}} \fi
The concept underlying the proposed saturation defense technique, as presented in \citep{Nayebi17}, aims to encourage neurons to operate within the saturated regions of their activation curves. As part of our approach to foster the emergence of AND-like neurons, we introduce a modified version of this idea as our third measure. We showed how the introduction of SDS allowed to characterise the AND-like nature of a neuron through the shapes of the $nics$ distributions when the neuron is fed with EDS and with SDS. Let us mimic the blue curve, depicted in Fig ~\ref{andorstat}, with the function $K(x) = |tanh(4*(x-0.5))|$. The comparison between the two distributions can be obtained by calculating the difference of the expected value of $K(x)|\mathrm{EDS}$ and the expected value of $K(x)|\mathrm{SDS}$; we call this quantity \textit{hyper-saturation (hypersat)}: 

\vskip 0.25cm
\begin{tabular}{p{14.6cm} c}
\sbo{$ hypersat = 
E(K(x)|\mathrm{EDS}) - 
E(K(x)|\mathrm{SDS}) $} &
\sbo{(5)}
\end{tabular}
\vskip 0.25cm

\if\afhead1 {\parok{rationale}} \fi
$hypersat$ tends to increase if the inputs of the neuron tend to co-occur, i.e. if they tend to take high values (and low values) simultaneously, more than expected by chance. The contrast between the real occurrence and the occurrence by chance is captured by the difference between the statistics calculated on EDS and SDS. This measure reminds of the saturation criterion imposed in \citep{Nayebi17}, with, however, a key difference. Thanks to the first measure (weight normalisation), the neuron's activation corresponds now to the neuron's $nics$: this renders the achievement of saturation much more difficult (but correspondingly more meaningful). 

\subsection*{Weight concentration} 

\if\afhead1 {\parok{weight concentration}} \fi
As indicated in section~\ref{section:trade-off}, achieving high $nics$ values, which indicate an AND-like nature of a neuron, is facilitated when the L1 norm of the neuron's weight vector is ``concentrated'' in a limited number of neurons. In other words, it is advantageous for only a few inputs to have high absolute weights, while the weights associated with other inputs remain close to zero. This tendency arises from the fact that a smaller number of inputs is more likely to exhibit correlation and co-occurrence. To encourage the concentration of weights, we multiply the gradient by the absolute value of the weight. Consequently, larger weights undergo more significant changes, and over time, the absolute values of the weights tend to diverge.

\subsection*{Feedback from classification task} 

\if\afhead1 {\parok{participation factor}} \fi
Not all neurons contribute equally to the supervised task (e.g., classification), which could render defense measures ineffective if implemented in non-relevant neurons. To address this issue, we calculate the ``participation factor'' for each neuron. This involves defining a loss as the mean absolute values of network outputs and computing the gradients for this loss. The gradient of each connection reflects its contribution to the output values. The participation factor of a neuron is then determined by summing the absolute values of the gradients associated with its outgoing connections.

\section{Experimental results}  

\if\afhead1 {\parok{ds, net architecture}} \fi
Tests have been conducted on the MNIST dataset \citep{LeCun98}. The network architecture used for our experiments is composed of 5 layers: 
\begin{itemize}
   \setlength\itemsep{-4pt}
   \item layer 0: input, 28x28 pixels  
   \item layer 1: convolutional, 12 filters, kernel size = 5, max pool 2x2
   \item layer 2: convolutional, 16 filters, kernel size = 5, max pool 2x2
   \item layer 3: linear, 120 neurons
   \item layer 4: linear, 84 neurons 
   \item layer 5: linear, 10 neurons
\end{itemize}

\if\afhead1 {\parok{ds, net architecture}} \fi
Training is carried out for a number (400) of epochs. At the beginning of the batch cycle, weight normalisation is performed according to equation (4). Subsequently, SDS is computed and two losses and the corresponding gradients are calculated: a classification loss based on cross entropy and a second loss, aimed at fostering the emergence of AND-like neurons. The gradient used to update the weights is a weighted average of the gradients obtained from these two losses. The pseudocode of the training cycle is reported hereafter.

\if\afhead1 {\parok{pseudocode}} \fi
\begin{Verbatim}[baselinestretch=1.0]      
For epoch = 0 To NofEpochs-1
   For batch = 0 To NofBatches-1
      L1 normalisation of neurons' weights
      generation of Scrambled Dataset
      calculation of classification loss
      calculation of corresponding gradient (1)
      calculation of loss from defence measures
      calculation of corresponding gradient (2)
      mix of gradients (1) and (2)
      gradient concentration
      update of weights
   Next batch
Next epoch
\end{Verbatim}

\if\afhead1 {\parok{loss2}} \fi
The second loss is based on the $hypersat_i$ factor described in the previous section. It is defined as:

\vskip 0.25cms
\begin{tabular}{p{14cm} c}
\newcommand{\ela}{hypersat_i \cdot partf_i}
\sbo{$ loss2 = 1-\dfrac {\sum_i (\ela)} {NofNeurons} $} &
\sbo{(6)}
\end{tabular}
\vskip 0.25cm

\noindent
$hypersat_i$ represents the hyper-saturation of neuron i, and the sum is carried out over all neurons. The term $partf_i$ represents the participation factor of neuron i. Multiplying $hypersat_i$ by this coefficient translates to giving a higher weight to neurons which play a more important role in classification. Without this term, loss2 could be dominated by neurons that do not participate in the signals used for classification. 

\if\afhead1 {\parok{neuron precision by layer}} \fi
The prevalence of AND-like neurons is expected to be higher in the initial layers of a neural network due to the need for a strong foundation in constructing a reliable structure. This design intuition suggests that the lower layers should possess a higher proportion of AND-like neurons, gradually decreasing in subsequent layers. By implementing this requirement, the network can maintain stability and ensure that the upper layers can be relatively weaker without jeopardizing the overall integrity of the construction.

\begin{table}[hb]
\center{
\begin{tabular}{ P{3cm} P{4cm} P{4cm} P{4cm}} \hline
epsilon & accuracy & accuracy & accuracy \\
 & (no attack) & (FGSM attack) & (PGD attack) \\ \hline
0.00 & 0.974   & 0.964   & 0.964   \\ 
0.05 & 0.774   & 0.890   & 0.833   \\ 
0.10 & 0.294   & 0.871   & 0.804   \\ 
0.15 & 0.083   & 0.852   & 0.762   \\ 
0.20 & 0.015   & 0.824   & 0.698   \\ 
0.25 & 0.002   & 0.792   & 0.591   \\ 
0.30 & 0.000   & 0.763   & 0.530   \\ \hline
\end{tabular}}
\caption{Classification results for the network under FGSM and PGD attacks, calculated on the MNIST test set (10000 examples) for 6 perturbation epsilon values.}
\label{accurt}
\end{table}

\begin{figure}[t] \begin{center} \hspace*{-0.0cm}
\includegraphics[width=17.00cm]{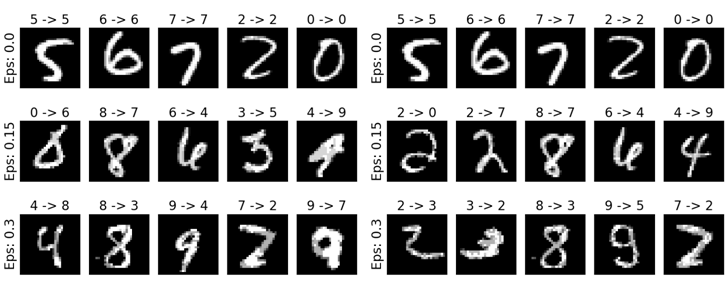}
\caption{
\if\afhead1 {\parok{figure}} \fi
\if\afhead1 {\parok{kaption}} \fi
Examples of digits correctly and incorrectly classified under FSGM attack (on the left) and PGD attack (on the right). The perturbations needed to cause incorrect classification are more ``semantic'', i.e. they are more aligned with human perception (this is particularly evident for the 8 $>$ 3 and 9 $>$ 4 transitions).}
\label{samples}
\end{center} \end{figure}

\begin{figure}[t] \begin{center} \hspace*{-0.5cm}
\includegraphics[width=18.00cm]{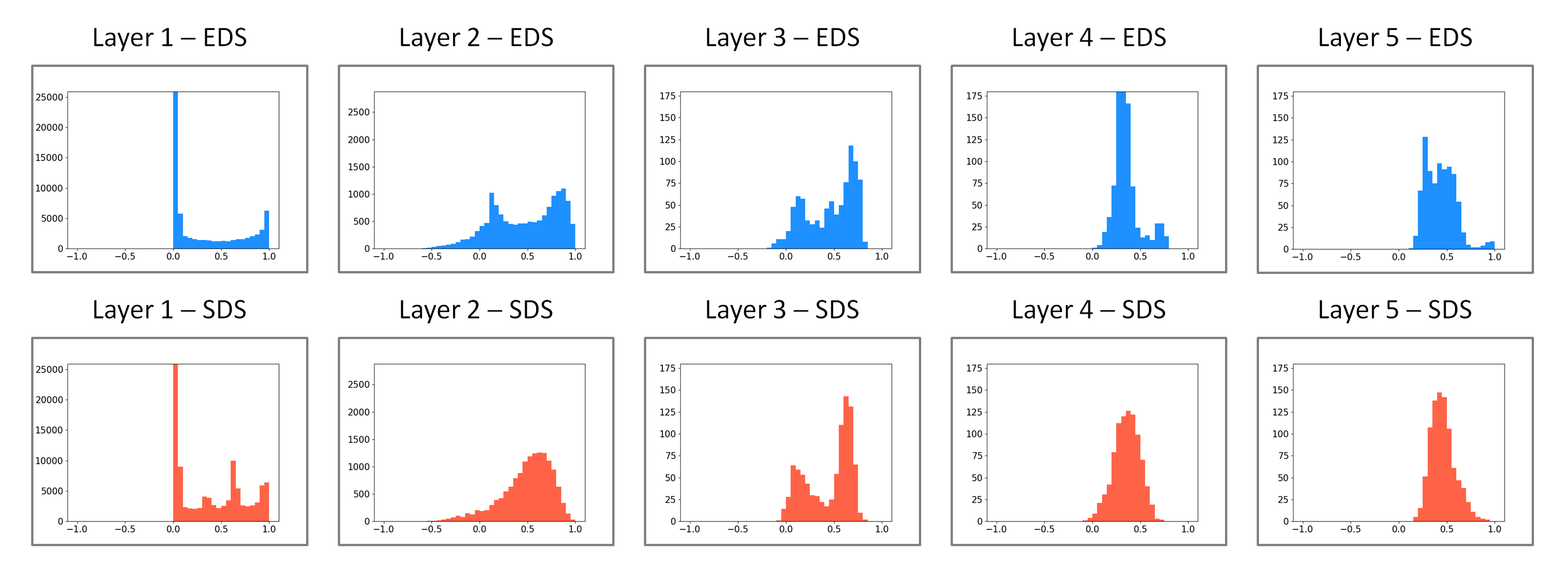}
\caption{
\if\afhead1 {\parok{figure}} \fi
\if\afhead1 {\parok{kaption}} \fi
The histograms depict the average distribution of $nics$ values for neurons in different layers after training. Neurons supplied with EDS (shown in blue) have bimodal histograms, while neurons supplied with SDS (orange histograms) show a tendency towards a normal distribution.}
\label{zistograms}
\end{center} \end{figure}

\if\afhead1 {\pardr{tests, results, samples}} \fi
The trained networks have been tested against FGSM and PGD attacks \citep{Kurakin16}, with different epsilon values: 0.05, ... 0.30. The experimental results (table.~\ref{accurt}) hint that the proposed method is effective in reducing the vulnerability of neural networks to adversarial perturbations. Fig.~\ref{samples} shows examples of digits correctly and incorrectly classified under FSGM and PGD attacks. The perturbations needed to cause incorrect classification are more ``semantic'', i.e. they are more aligned with human perception. Fig.~\ref{zistograms} shows histograms of $nics$ of typical neurons for each layer. Neurons supplied with EDS have bimodal histograms, while neurons supplied with SDS show a tendency towards a normal distribution.  

\section{Discussion} 

\if\afhead1 {\parok{log-stats, brain}} \fi
In this work we propose the idea that AND-like neurons are helpful to combat adversarial examples. Since neural networks take inspiration from biology, it is worth asking if this idea has some biological underpinnings. Contrary to what it is often assumed, the variables of functional and structural brain parameters (e.g. the synaptic weights, the firing rates of individual neurons, the synchronous discharge of neural populations, the number and size of synaptic contacts between neurons) do not have a bell-shaped distribution. In fact, the distribution of many such parameters is strongly skewed with a heavy tail, suggesting that a lognormal distribution is a better fit for experimental data \citep{Buzsaki14}. 

\if\afhead1 {\parok{log-stats, consistency with model}} \fi
These findings are compatible with the presence or prevalence of AND-like operations in biological neurons, whose firing rates tend to be simultaneously high or simultaneously low. We should also consider that communication between biological neurons happens by means of standard spikes, whereby the signal is encoded in the spike frequency. It is possible that an additive rule in the frequency domain (relevant to biological neurons) translates to a multiplicative rule in the continuous domain (relevant to artificial neurons). The use of spiking neural networks might be an option to close the gap between biological reality and machine learning models.

\if\afhead1 {\parok{spine size, brain and consistency with model}} \fi
As already pointed out, AND-like operations can be more easily achieved when the number of significant inputs is low. This corresponds to a situation in which most synaptic weights are low and only a handful are high enough to make significant contributions to the computation. As reported in \citep{Loewenstein11}, the stationary distribution of spine sizes of neurons is consistent with a log-normal function. This corresponds to a situation in which most spines are small, and a small subset are much larger: a finding consistent with our model. 

\if\afhead1 {\parok{spine size change, brain and consistency with model}} \fi
Spine sizes show significant fluctuations that can occur over periods ranging from days to months. The extent of changes in spine sizes appears to be proportional to the size of the spine, as noted in previous research \citep{Loewenstein11}. This multiplicative dynamic differs from the conventional additive dynamics assumed in traditional neuron models. These findings align with our proposed model of neuron operation, specifically the gradient update rule that involves multiplying the gradient by the absolute value of the weight, which might correspond to spine size in biological systems.

\if\afhead1 {\parok{SDS captures andness}} \fi
The comparison between EDS and SDS effectively captures the concept of AND-like behaviour in neurons. When the inputs of a neuron are correlated and tend to have both high and low values together, the statistical behaviour of the neuron's input set becomes more AND-like. Moreover, if there are more correlated inputs, the neuron exhibits a stronger AND-like behaviour. For example, a neuron with 100 correlated inputs is more AND-like compared to a neuron with only 10 correlated inputs because the likelihood of having a higher number of correlated inputs is lower. This relationship is visually depicted by the shape of the probability distribution conditioned on SDS, where the standard deviation becomes smaller as the number of correlated inputs (N) increases.

\if\afhead1 {\parok{human intuition}} \fi
Perhaps AND-like neurons align more closely with human intuition. For instance, to say it is Christmas (in the northern hemisphere), several conditions need to be satisfied simultaneously, such as winter season, Santa Claus puppets on house walls, cold weather, and reuniting with friends and relatives. Making decisions, like buying gifts, requires complete confidence that it is indeed Christmas. Knowing with only 70\% confidence that it is Christmas, based on, for example, 60 out of 90 input features being true, is less useful. However, networks dominated by perfect AND gates are not practical because they would respond only to specific combinations of input values, limiting pattern emergence and reducing classification capability. Similarly, networks dominated by OR neurons are also not beneficial. Instead, a balanced combination of AND and OR neurons is optimal, as the Latin saying ``in medio stat virtus'' suggests.
 
\if\afhead1 {\parok{interpretability}} \fi
The AND-like character of neurons can be also linked to the notion of interpretability. Concept definitions are essentially obtained by means of AND clauses. A lemon is a yellow fruit which grows in Spain; this definition can be formalised as a series of AND operators: ``lemon'' = True iff ``fruit'' = True AND ``yellow'' = True AND ``grows in Spain'' = True. There might be space for some OR operator as well (e.g.: a lemon is a yellow fruit which grows in Spain OR in Sicily), but only to a limited extent). We could push this argument as far as to say that the objective of learning is essentially that of turning data framed with OR clauses (such is the nature of a dataset) into data framed \textit{more} with AND clauses (this is the nature of the features encoded in higher-level neurons).  

\section{Conclusion} 

\if\afhead1 {\parok{conclusion}} \fi
In this work, we have introduced a novel approach to tackle the challenge of adversarial examples. Our primary objective is to enhance the network's ability to withstand such adversarial attacks by augmenting the proportion of neurons that exhibit AND-like behaviour, through a set measures. By employing these techniques, we have conducted experiments on the popular MNIST dataset, subjecting it to FSGM and PGD attacks. The outcomes of these experiments demonstrate the potential effectiveness of our proposed method in bolstering adversarial robustness. Nonetheless, additional experiments are imperative to validate the applicability and efficacy of this approach when applied to different models, datasets, and types of attacks.

\bibliographystyle{apalike}
\bibliography{laidadefn}
 
\end{document}